%% file: tccml_iclr2025_conference.tex
\documentclass{article} % For LaTeX2e
\usepackage{tccml_iclr2025_conference,times}

\input{math_commands.tex}

\usepackage{graphicx}
\usepackage{hyperref}
\usepackage{url}
\usepackage{booktabs}
\usepackage{float}

\title{XAI4Extremes: An interpretable machine learning framework for understanding extreme-weather precursors under climate change}

\author{{Jiawen Wei} \\
National University \\
of Singapore \\
\And
{Aniruddha Bora \& Vivek Oommen} \\
Brown University \\
\And
{Chenyu Dong} \\
National University \\
of Singapore \\
\And
{Juntao Yang \& Jeff Adie} \\
NVIDIA AI \\
Technology Centre \\
\And
{Chen Chen} \\
Centre for Climate \\
Research Singapore\\
\And
{Simon See} \\
NVIDIA AI \\
Technology Centre \\ 
\And
{George Karniadakis} \\
Brown University \\
\texttt{george\_karniadakis@brown.edu}
\And
{Gianmarco Mengaldo} \\
National University of Singapore \\
\texttt{mpegim@nus.edu.sg}
}

\iclrfinalcopy % Uncomment for camera-ready version, but NOT for submission.
\begin{document}

\maketitle

\begin{abstract}
Extreme weather events are increasing in frequency and intensity due to climate change. 
This, in turn, is exacting a significant toll in communities worldwide. 
While prediction skills are increasing with advances in numerical weather prediction and artificial intelligence tools, extreme weather still present challenges.
More specifically, identifying the precursors of such extreme weather events and how these precursors may evolve under climate change remain unclear. 
In this paper, we propose to use post-hoc interpretability methods to construct relevance weather maps that show the key extreme-weather precursors identified by deep learning models.
We then compare this machine view with existing domain knowledge to understand whether deep learning models identified patterns in data that may enrich our understanding of extreme-weather precursors. 
We finally bin these relevant maps into different multi-year time periods to understand the role that climate change is having on these precursors.
The experiments are carried out on Indochina heatwaves, but the methodology can be readily extended to other extreme weather events worldwide.
\end{abstract}

\section{Introduction}
\label{sec:introduction}
Climate change is playing a pivotal role in exacerbating extreme weather across the globe, with extreme weather events becoming more frequent and severe~\citep{masson2021climate}. 
These events, in turn, are exerting heavy socioeconomic and environmental tolls on communities and fragile ecosystems worldwide. 
For instance, the tropical Indo-Pacific is witnessing an increased frequency of heatwaves and extreme precipitation due to critical changes in synoptic weather patterns in the region~\citep{dong2024indo}.
Similarly, heatwaves during the summer and storms during the winter are becoming more frequent in Europe due to atmospheric circulation changes~\citep{faranda2023atmospheric}. 
Other regions are also experiencing an increased frequency of heatwaves and other extremes -- see for instance~\citep{perkins2020increasing,donat2016more}.

Extreme weather is commonly defined as weather that is significantly different from the typical conditions for a particular region and time of the year.  
Heatwaves, for example, are due to prolonged periods of excessively hot weather relative to the expected conditions in a given area and time, that can lead to significant impacts on the affected community, such as health, infrastructure, and agricultural issues. In order to mitigate the impact of weather extremes, authorities typically issue warnings ahead of time to alert the population that may be at risk.
These warnings are largely based on weather forecasts provided by local and and global operational weather services. 
However, the prediction skills for certain types of extremes and regions is still relatively poor, for both short-range (a few days ahead) and mid-range (two weeks ahead) forecasts.
One example is related to the forecast of heatwaves in large portions of the tropics, including the Maritime continent, and central Africa, as well as in the Carribean and Central America and the western US~\citep{de2018global}.
 
Indeed, predictability drivers for heatwaves vary across different regions, and they are due to the confluence of several physical mechanisms, including diabatic heating from radiation and surface heat fluxes, adiabatic warming from air subsidence, and horizontal movement of hot air masses~\citep{de2018global}. 
At mid-latitudes, heatwaves are frequently associated with persistent atmospheric blocking events, which promote subsidence and clear-sky conditions, thereby enhancing surface warming~\citep{kautz2022atmospheric}.
In the tropics, typical drivers are instead dominated by high solar radiation and reduced cloud cover that amplify surface heating, as well as by suppressed convection from large-scale subsidence, and warm sea surface temperature anomalies during e.g., El Niño events~\citep{cai2014increasing}.
In addition, land conditions and vegetation, such as land moisture levels and presence of forest vs grassland are important drivers of heatwaves~\cite{domeisen2023prediction}. 
These drivers help increasing the confidence in the forecast of heatwaves, providing an important time window to issue early warnings to the affected population.
These drivers, also referred to as precursors, are typically the result of human-expert knowledge, or briefly the ``\textit{human view}''. 
In this work, we take a different perspective, and look at these precursors through the lenses of interpretable machine learning (ML), thereby providing a possibly complementary ``\textit{machine view}''. The latter is obtained by identifying what data the machine deemed important to the onset of heatwaves, and it is used to understand (by working with human domain experts) whether it may be helpful in enriching our understanding of precursors -- see also~\cite{mengaldo2024explain} for the use of explainable artificial intelligence (XAI) for scientific knowledge discovery.
Without losing generality in the methodology proposed, we focus on tropical heatwaves in the Indochina peninsula, and attempt to answer two questions via interpretable ML: (i) What are the key precursors of these events? (ii) Is climate change influencing these precursors?

% This work can help in two different directions. 
% On the one hand, it can provide a complementary view on extreme weather precursors, that may diverge from existing human-expert knowledge, thereby offering a pathway for knowledge discovery. 
% On the other hand, by identifying the input data at the granular level -- i.e., granular features -- that were deemed important by the machine, it can help shaping ML model behavior and improve ML predictive accuracy (e.g., by synthesizing relevant samples for adversarial training). In addition, with the advent of several AI solution for weather and climate science~\citep{pathak2022fourcastnet,lam2023learning,bodnar2024aurora,nguyen2023climax,wang2024orbit}, we believe that interpretable ML may prove important to both understand and shape model behaviour. 
% The latter is especially critical for tasks where AI may struggle to generalize to future out-of-sample scenarios, such as in the context of climate change. 
% Furthermore, better understanding the precursors to weather extremes, and how these may be changing under our evolving climate might prove crucial for increasing the forecasting skills of numerical weather prediction systems, whether they are equation-~\citep[e.g.,][]{wedi2015modelling} or AI-based~\citep[e.g.,][]{lang2024aifs}. 

\section{Methodology}
\label{sec:methodology}
To outline our approach, we focus on heatwaves in the Indochina peninsula (the latter depicted in Figure~\ref{fig:mask_indochina}). 
These can be divided into heatwaves in the dry and in the wet seasons, whereby the precursors and onset mechanisms differ~\citep{luo2018synoptic}. 
We focus on dry-season (FMAM) heatwaves without lacking generality on the methodology proposed here. 

The key idea is to look at these extreme weather events, namely dry-season heatwaves, using interpretable machine learning; more specifically post-hoc interpretability methods applied to a binary time series classification deep learning (DL) framework. 
This approach allows producing relevance maps, that highlight what input data the DL framework deemed important for the prediction it made. 
The binary DL time series classification framework is setup as follows. 
As input data, we consider the spatial (i.e., geographical) maps of 23 variables for the 7 days prior of a heatwave striking the Indochina peninsula.
The 23 input variables characterize the large majority of dry-season heatwave precursors, and the 7 days time window provides a relevant time frame to capture the underlying pathways leading to these extremes.
We then assume that the DL framework is able to identify patterns in the data that are causal to heatwaves; in other words, we assume that it could capture systematically the precursors to heatwaves. 
Indeed, we consider only true positive samples, such that the data deemed important by the DL framework, also referred to as ``\textit{machine view}'', is only associated to correctly classified heatwaves. 
The binary labels for the classification task are (1) heatwave and (0) non-heatwave, where the heatwaves are identified as outlined in Appendix~\ref{app:heatwaves}.

\begin{figure}[htpb]
    \centering
    \includegraphics[width=1.0\linewidth]{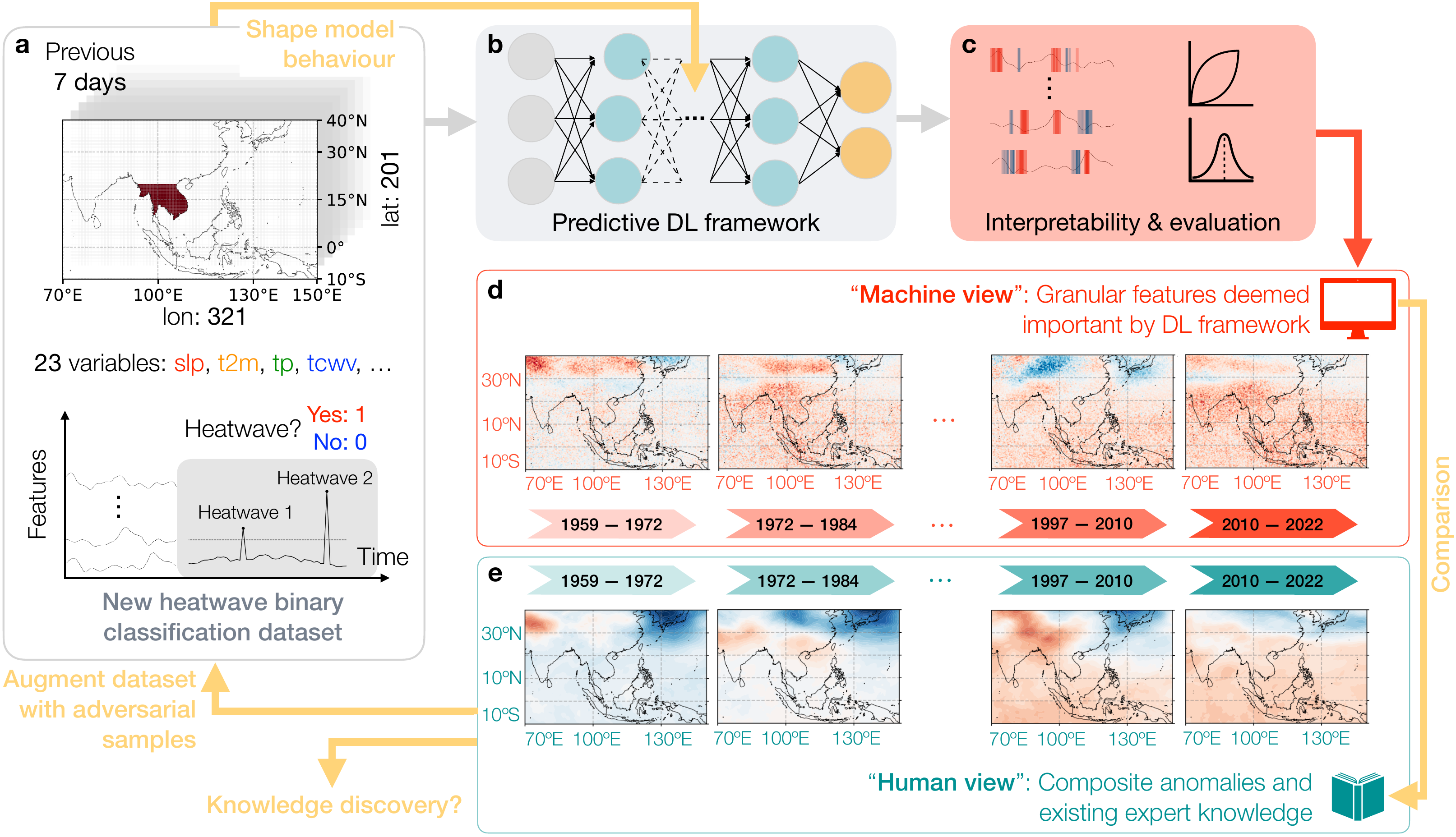}
    \caption{The XAI4Extremes framework proposed, composed of a novel extreme weather dataset (a), a DL predictive model (b), an interpretability block along with its evaluation (c), that produces relevance maps, or what we called the ``\textit{machine view}'' (d). The latter (d) is then compared with existing human expert knowledge (e) for knowledge discovery or for augmenting the dataset with e.g., adversarial samples that can shape and improve model behavior.}
    \label{fig:method}
\end{figure}

The final heatwave binary classification dataset consists of 720 samples with an approximate ratio of (1) heatwave vs (0) non-heatwave being 1:5.
We split the dataset into training, validation, and testing sets with a ratio of [0.6:0.2:0.2], and then train the Transformer model for heatwave classification. 
We apply four different post-hoc interpretability methods, namely Integrated Gradients~\citep{sundararajan2017axiomatic}, DeepLIFT~\citep{shrikumar2017learning}, DeepSHAP~\citep{lundberg2017unified}, and GradSHAP~\citep{lundberg2017unified}, to the trained Transformer model. 
To guarantee that we obtain the most accurate and robust relevance maps, we adopt the interpretability evaluation frameworks in \cite{turbe2023evaluation} and \cite{wei2024revisiting}. 
Integrated Gradients performs the best among four post-hoc methods according to the evaluation results; thereby we use the relevance maps it generates for analysis. 
The overall approach, that we name XAI4Extremes, is depicted in Figure~\ref{fig:method}: we propose a new dataset for weather extremes -- heatwaves in this particular case (panel a, in gray), that is used by a predictive DL framework (panel b, in blue), to which we apply post-hoc interpretability and its evaluation (panel c, in red). 
The relevance maps produced by the post-hoc interpretability method, what we also refer to as ``\textit{machine view}'' (panel d, in red), are then compared against human expert knowledge, what we also refer to as ``\textit{human view}'' (panel e, in green). 
This comparison may lead to knowledge discovery in terms of heatwave precursors and role of climate change in heatwave precursors. 
This may be the case when the machine view enriches human expert knowledge, by providing a scientifically plausible use of data that was unknown to human domain experts, but that domain experts can explain.
Indeed, it is responsibility of human domain experts to respond to the question \textbf{why} the interpretable machine learning framework deemed important a specific set of input data.
The relevance maps can also be used to generate adversarial samples to augment the dataset and shape model behavior, thereby improving the performance of the predictive DL framework.
We remark that the approach outlined in this section can readily be applied to other types of weather extremes in different regions worldwide.

\section{Results}
\label{sec:results}
\begin{figure}[htpb]
    \centering
    \includegraphics[width=1.0\linewidth]{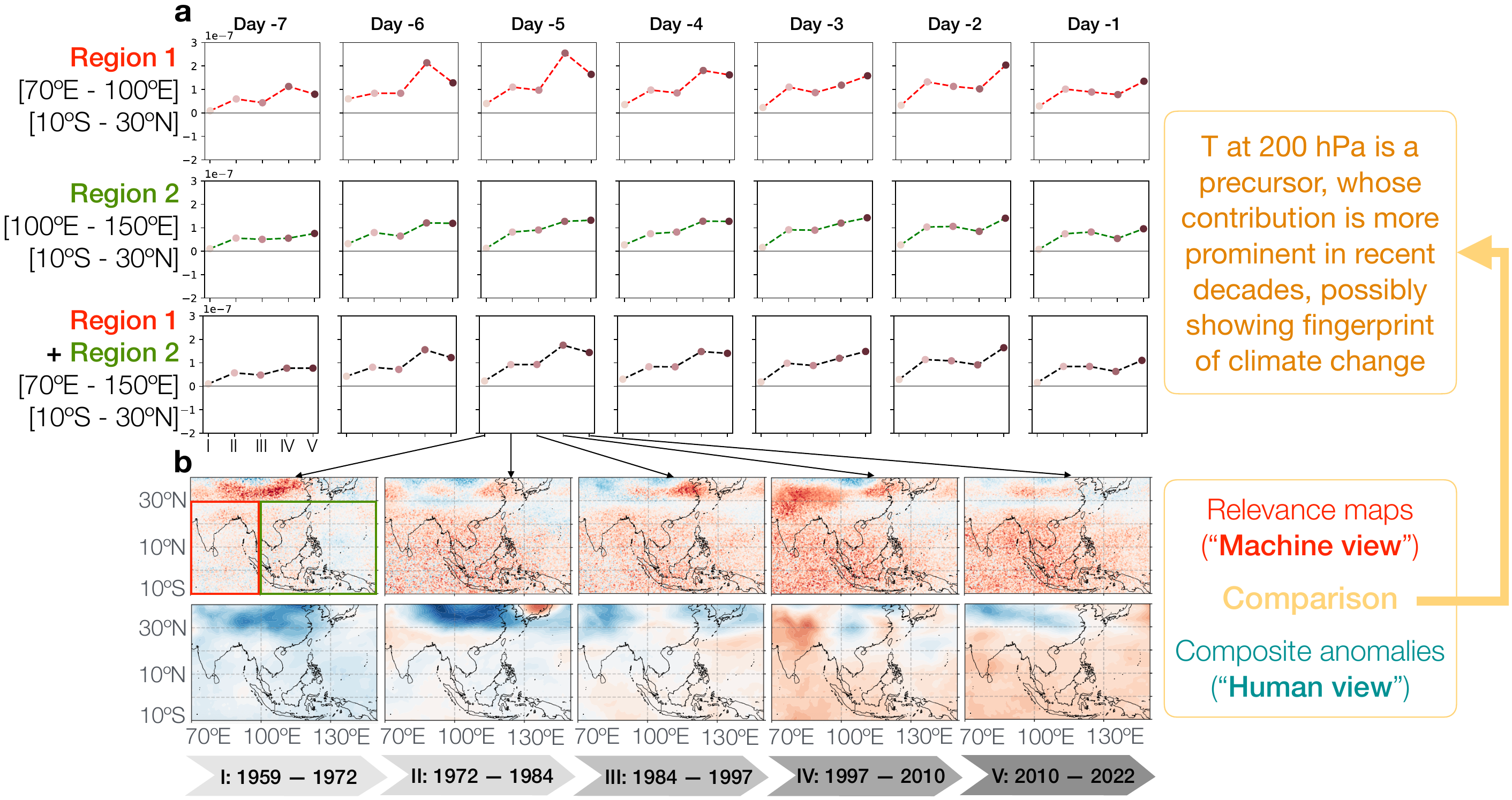}
    \caption{Mean relevance of temperature at 200 hPa for the 5 historical time periods considered and on the 7 days prior to heatwaves in Indochina, for region 1 (a, top), region 2 (a, middle), region 1+2 (a, bottom), along with the corresponding relevance maps -- i.e., ``\textit{machine view}'' -- associated to 7 days prior to heatwave, and composite anomalies -- i.e., ``\textit{human view}'' -- (b).}
    \label{fig:results}
\end{figure}

Figure~\ref{fig:results} shows the temperature field at 200 hPa, that is the temperature between approximately 11 and 12 km altitude (i.e., the temperature in the upper troposphere), for two different regions, region 1 and 2. 
Region 1 comprises the Indian Ocean, and India, while region 2 comprises the Maritime continent and part of the Pacific Ocean. 
In Figure~\ref{fig:results}, panel a, we show the mean trend of relevance for region 1 (top row), region 2 (middle row), and region 1 and 2 combined (bottom row). 
It is possible to see how the temperature in the upper troposphere is deemed more important by the machine for heatwaves in Indochina in more recent decades for both regions, with a clear upward trend. 
If we compare the interpretability results (i.e., the relevance maps or machine view) with something more understandable by humans, i.e., composite anomalies, we note that there is indeed a warming of the upper troposphere that is associated to heatwaves in Indochina (Figure~\ref{fig:results}, panel b).
This indicates that the temperature at 200 hPa is becoming a key precursor of Indochina heatwaves, especially in recent decades (in agreement with composite anomalies -- see Figure~\ref{fig:rel_ano_t200hPa} in Appendix~\ref{app:comparison}), aspect that may indicate the fingerprint of climate change. 
If we further compare the contribution of the other 22 variables, we observe how the temperature in the upper troposphere is one of the most important variables for the DL prediction -- see Appendix~\ref{app:mean-relevance}, i.e., Figures~\ref{fig:relevance_region1} to \ref{fig:relevance_pos_region12}. 
In addition, if we look at the contribution of two common variables related to climate warming, namely 2-meter temperature and max temperature (Appendix~\ref{app:comparison}, Figures~\ref{fig:rel_ano_t2m} and ~\ref{fig:rel_ano_txm}, respectively), we observe how these have a strong trend in terms of composite anomalies, but not in terms of relevance maps. 
This result points to a human-understandable explanation where higher 200 hPa temperature can suppress convection and increase subsidence, thereby leading reduced cloud cover that amplifies surface heating, potentially leading to heatwaves. Onset mechanism that may have been exacerbated by climate change, with a fingerprint on the 200 hPa temperature.
In Appendix~\ref{app:predictive-performance}, Table~\ref{tab:prediction-performance}, we also show the predictive performance of the DL model for the classification task, on the 5 historical time periods considered. 
We observe that the performance can be significantly improved, aspect that is currently ongoing, with some preliminary results provided in Table~\ref{tab:other-models}.

\section{Conclusions}
\label{sec:conclusions}

The preliminary results presented in this work aim to introduce the overarching explainable AI framework we propose, namely XAI4Extremes.
The framework has the key objective of better understanding weather extremes and their evolution under climate change.
To achieve this task, we create a novel binary classification dataset for heatwaves in Indochina (noting that the same dataset type can be created for other weather extremes in other regions worldwide).
We then propose to couple a predictive DL framework with interpretability methods, in order to understand \textit{\textbf{what}} data the machine deemed important for its predictive performance of true positive samples (i.e., correctly identified heatwaves), something we refer to as ``\textit{machine view}''.
We finally propose to compare this machine view to existing human expert knowledge (what we call ``\textit{human view}''), to respond the question \textit{\textbf{why}} the machine used those data. 
The latter aspect may lead to knowledge discovery, or it can be used to shape model behavior by e.g., generating ad-hoc adversarial samples based on the machine view. 
We note that there are still several, yet stimulating, open challenges to be overcome.
For instance, how to further improve the robustness of post-hoc interpretability methods, how to develop effective ante-hoc (also referred to as self-explainable) ML approaches~\citep[e.g.,][]{turbe2024protos} for spatio-temporal data,  
how to use relevance maps for spatio-temporal data, and isolate features that are relevant, how to guarantee that the patterns identified by the relevance maps are causal to the task, among others. 
We believe that these limitations are open opportunities for the AI and broader scientific research communities that can be tackled over the next few years.

\subsubsection*{Acknowledgments}
J.W.\ and G.M.\ acknowledge support from MOE Tier 2 grant T2EP50221-0017, and from MOE Tier 1 grant 22-4900-A0001-0.

\bibliography{iclr2025_conference}
\bibliographystyle{iclr2025_conference}

\appendix

\section{Identification of heatwaves}
\label{app:heatwaves}

Identifying heatwaves remains a significant challenge. 
Currently, there are numerous definitions of heatwaves in the research community, yet there is no consensus on a standard definition. 
This complexity arises from the varied spatial coverage and duration of heatwaves. 
In our study, we adopted a relatively simple two-stage definition that combines index-based and event-based approaches, which have been widely used in other research. 

We first define heatwaves on each individual grid point in the daily ERA5 reanalysis data from 1959 to 2022 using the heatwave index TX90pct~\citep{perkins2012increasing}. 
The threshold for one day at one grid point is the calendar day 90th percentile of the daily maximum temperature, based on a centered 15-day window. 
A heatwave is defined as three or more consecutive days exceeding this threshold, and all days belonging to this heatwave are considered as heatwave days for that grid point. 
We note that we removed a grid point by grid point linear trend from the the data. 
This is because we want to maintain a relatively uniform distribution of heatwaves in the studied period.

\begin{figure}[htpb]
    \centering
    \includegraphics[width=0.4\linewidth]{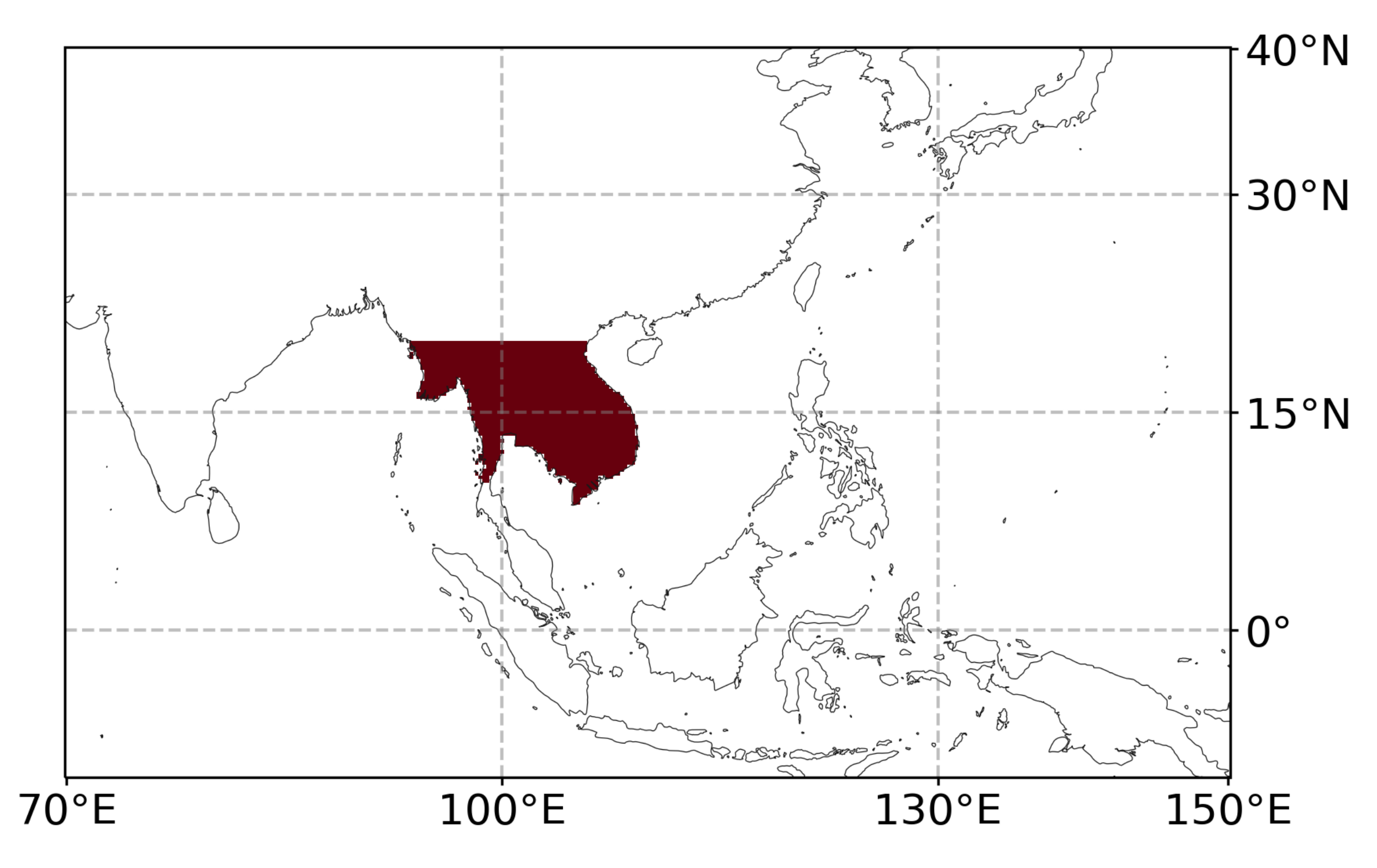}
    \caption{Indochina region used to define heatwaves (dark red).}
    \label{fig:mask_indochina}
\end{figure}
Based on this grid point by grid point definition of local heatwaves, we further define heatwave events in Indochina using the regional mask illustrated in figure~\ref{fig:mask_indochina}. 
For each region, one heatwave event is defined when a minimum number of grid points are identified as heatwaves. 
Specifically, this threshold is set at the 90th percentile of the number of grid points classified as heatwaves during the season of interest. 
We define the first day that exceeds the predefined threshold as the heatwave onset day. To avoid overlapping events, we stipulate that no day within the seven days preceding any onset day should exceed this threshold. 
For the onset days of non-extreme events, we randomly select days when the number of grid points falls below the predefined threshold within the same season, following specific criteria: We ensure that there are no heatwave onset days or other non-extreme events within a 7-day window before and after these selected days. 
We provide the dataset with a ratio of non-extreme events to extreme events set at 5:1. 
This is the maximum ratio achievable while following the selection strategy outlined above.

\section{Additional results}
\label{app:additional-results}

\subsection{Predictive performance of DL framework}
\label{app:predictive-performance}

In Table~\ref{tab:prediction-performance}, we show the predictive performance of the Transformer model used as a reference for this study. 
We observe that the performance can be significantly improved, something that is currently ongoing. 
To this end, in Table~\ref{tab:other-models}, we also present the results (for the testing set only) for Convolutional Encoder with Attention blocks (Conv+Attn) and FourCastNet~\citep{pathak2022fourcastnet} models that are being rolled out. 
The Conv+Attn model used in this study consists of three sets of convolution and maxpool operations (that down-samples the latent representation by a factor of 2) resulting in a latent representation at 1/8th of the input resolution. The first convolution block contains a channel-wise attention layer and the last convolution block is followed by a spatial attention block. The output of the last convolutional block is projected to a scalar using a fully connected layer with sigmoid activation, that predicts if whether or not the heat wave is going to occur. 
The FourCastNet used here is modified from the original code \citep{pathak2022fourcastnet} by adding a extra block specifically for the binary classification of the heatwaves.
\begin{table}[htpb]
    \centering
    \caption{Predictive performance of trained Transformer model}
    \begin{tabular}{cccccc}
    \toprule
     & Sensitivity & Specificity & Precision & Miss Rate & Accuracy \\
     & (TPR) & (TNR) & (PPV) & (FNR) &  \\
     \midrule
     % 27 Jan 1959 — 2 Feb 1972 & 52.38\% & 100.00\% & 100.00\% & 47.62\% & 93.06\% \\
     % 10 Feb 1972 — 4 Apr 1984 & 35.48\% & 100.00\% & 100.00\% & 64.52\% & 86.11\% \\
     % 16 Dec 1984 — 26 Feb 1997 & 57.69\% & 99.15\% & 93.75\% & 42.31\% & 91.67\% \\
     % 9 Mar 1997 — 16 Feb 2010 & 50.00\% & 100.00\% & 100.00\% & 50.00\% & 93.75\% \\
     % 26 Feb 2010 — 18 Dec 2022 & 66.67\% & 99.17\% & 94.12\% & 33.33\% & 93.75\% \\
     1959 — 1972 & 52.38\% & 100.00\% & 100.00\% & 47.62\% & 93.06\% \\
     1972 — 1984 & 35.48\% & 100.00\% & 100.00\% & 64.52\% & 86.11\% \\
     1984 — 1997 & 57.69\% & 99.15\% & 93.75\% & 42.31\% & 91.67\% \\
     1997 — 2010 & 50.00\% & 100.00\% & 100.00\% & 50.00\% & 93.75\% \\
     2010 — 2022 & 66.67\% & 99.17\% & 94.12\% & 33.33\% & 93.75\% \\
     \bottomrule
    \end{tabular}
    \label{tab:prediction-performance}
\end{table}
\begin{table}[htpb]
    \centering
    \caption{Predictive performance of other models being rolled out -- results shown are for the testing period only, that is between 26 Feb 2010 to 18 Dec 2022.}
    \begin{tabular}{cccccc}
    \toprule
     & Sensitivity & Specificity & Precision & Miss Rate & Accuracy \\
     & (TPR) & (TNR) & (PPV) & (FNR) &  \\
     \midrule
     Conv+Attn & 70.83\% & 96.67\% & 80.95\% & 29.16\% & 92.36\% \\
     FourCastNet & 70.83\% & 97.50\% & 85.00\% & 29.16\% & 93.75\% \\
     % Model 3 & 57.69\% & 99.15\% & 93.75\% & 42.31\% & 91.67\% \\
     % Model 4 & 50.00\% & 100.00\% & 100.00\% & 50.00\% & 93.75\% \\
     % Transformer & 66.67\% & 99.17\% & 94.12\% & 33.33\% & 93.75\% \\
     \bottomrule
    \end{tabular}
    \label{tab:other-models}
\end{table}

\subsection{Mean relevance for all variables}
\label{app:mean-relevance}

\begin{figure}
    \centering
    \includegraphics[width=1.0\linewidth]{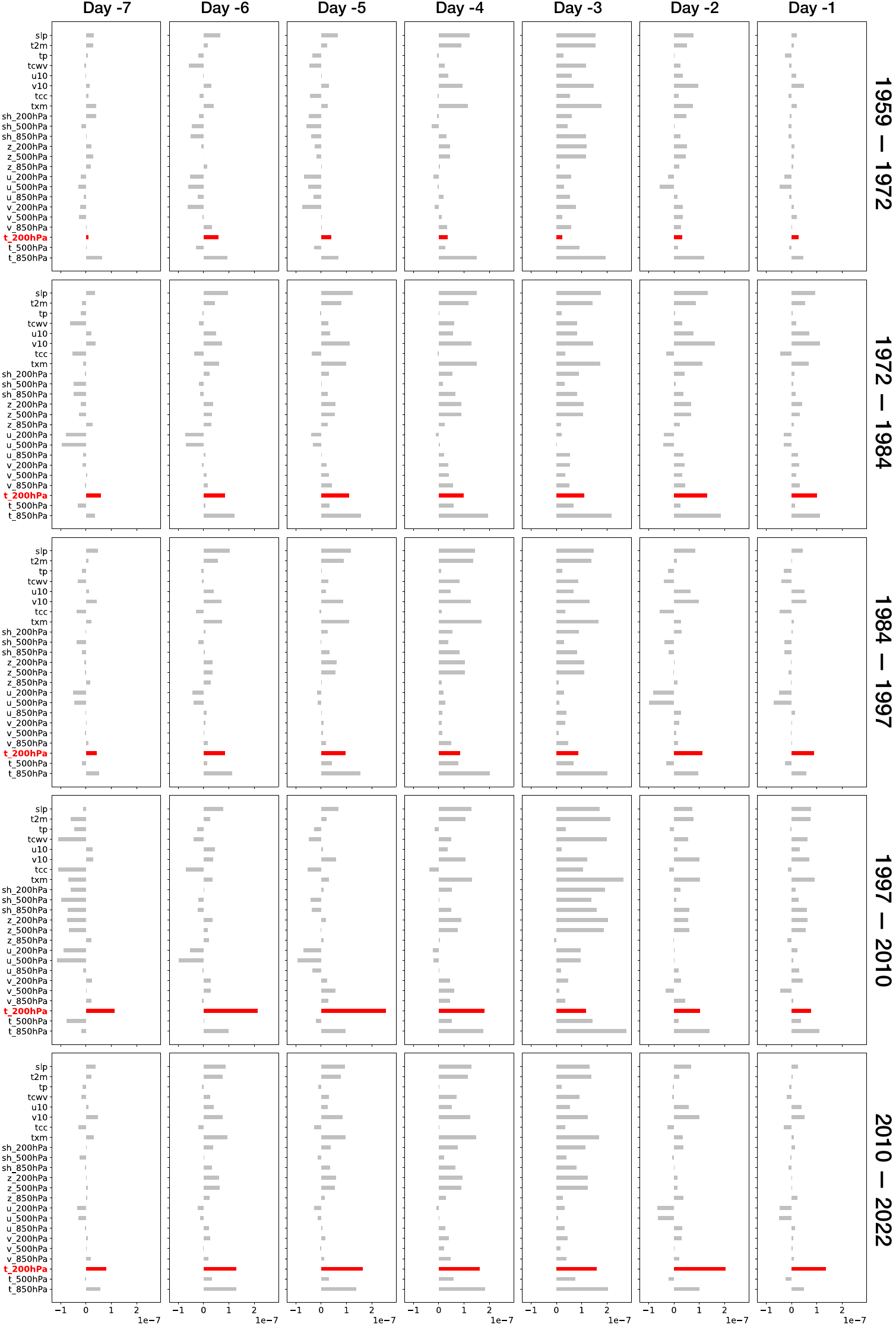}
    \caption{Mean relevance, \textbf{positive and negative}, on \textbf{region 1}, for all 23 variables considered, on the 7 days prior to heatwaves in Indochina and for the 5 historical time periods considered.}
    \label{fig:relevance_region1}
\end{figure}

\begin{figure}
    \centering
    \includegraphics[width=1\linewidth]{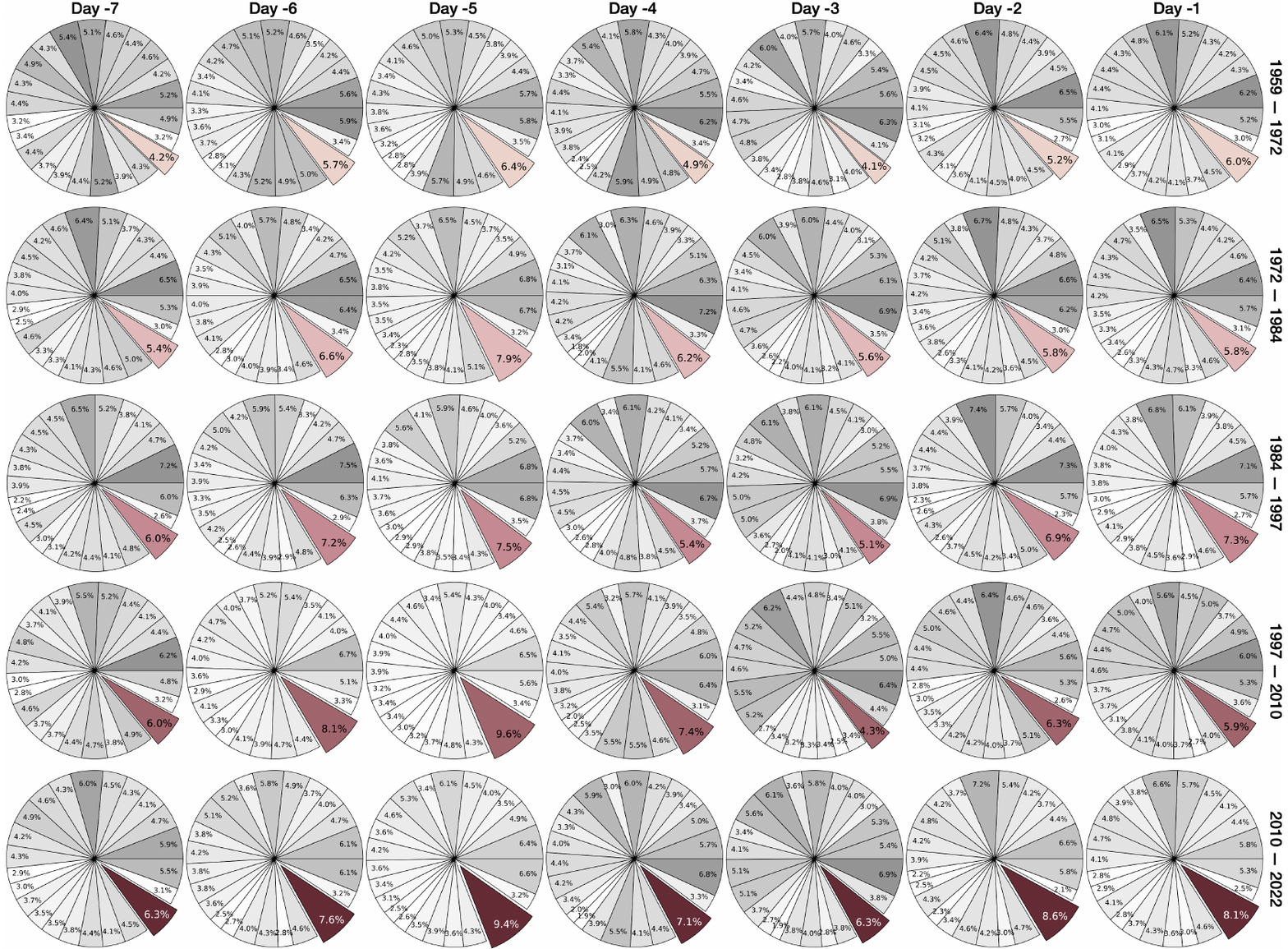}
    \caption{Mean relevance, \textbf{only positive} (negative is set to zero) on \textbf{region 1}, for all 23 variables considered, on the 7 days prior to heatwaves in Indochina and for the 5 historical time periods considered. Variable t\_200hPa (temperature at 200hPa) is highlighted with red colors across 5 historical time periods, while the other variables are displayed in gray.}
    \label{fig:relevance_pos_region1}
\end{figure}

\begin{figure}
    \centering
    \includegraphics[width=1.0\linewidth]{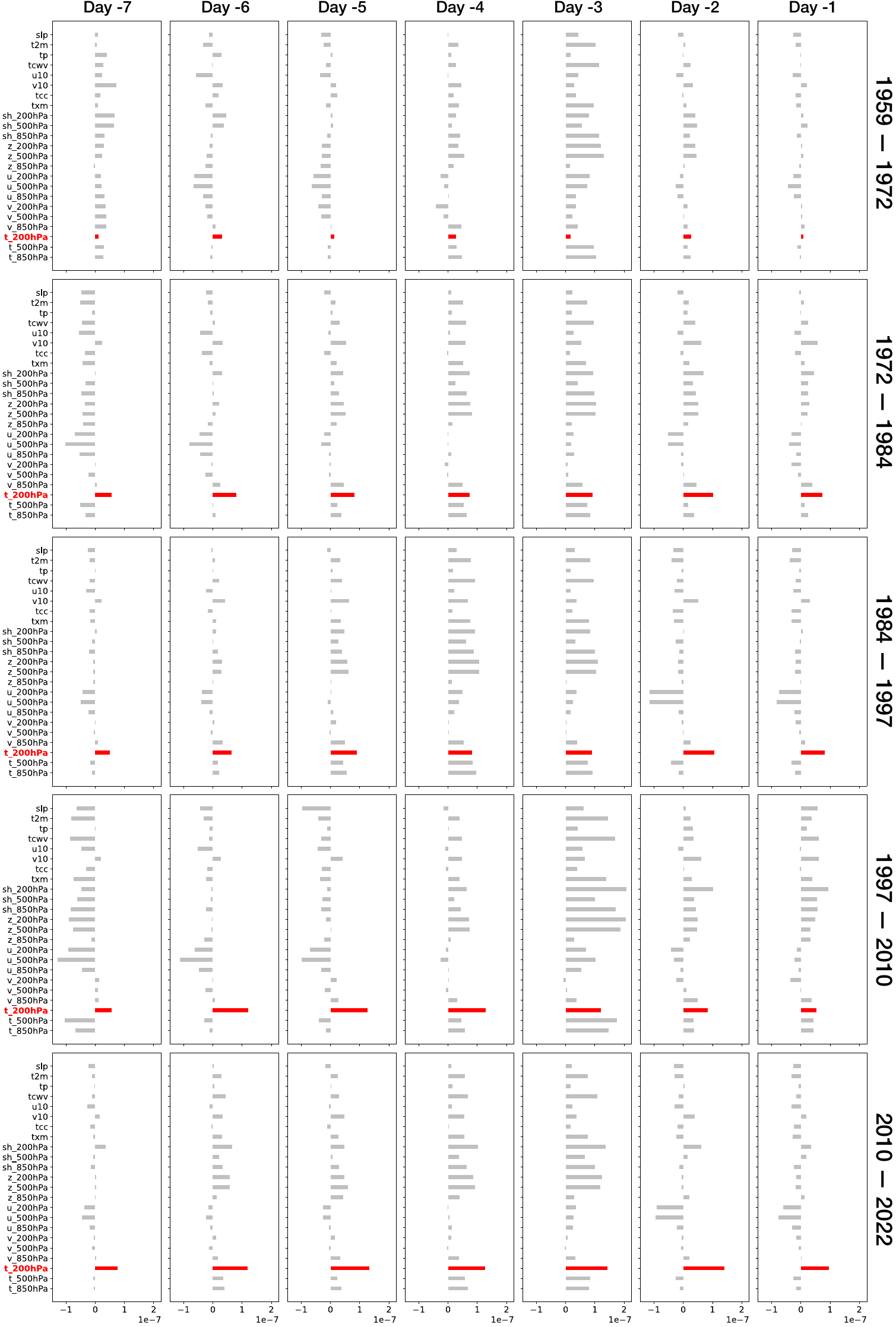}
    \caption{Mean relevance, \textbf{positive and negative}, on \textbf{region 2}, for all 23 variables considered, on the 7 days prior to heatwaves in Indochina and for the 5 historical time periods considered.}
    \label{fig:relevance_region2}
\end{figure}

\begin{figure}
    \centering
    \includegraphics[width=1.0\linewidth]{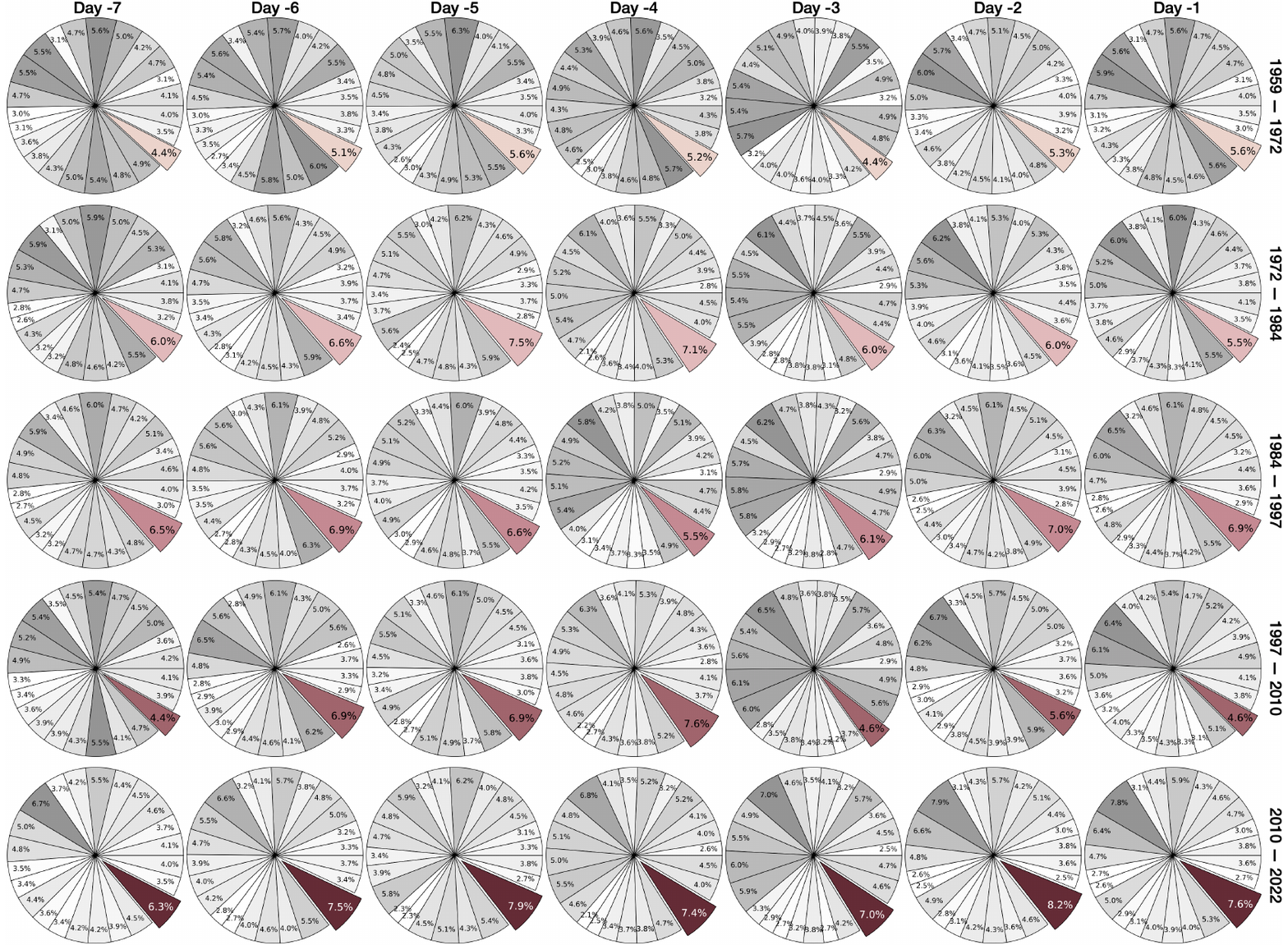}
    \caption{Mean relevance, \textbf{only positive} (negative is set to zero) on \textbf{region 2}, for all 23 variables considered, on the 7 days prior to heatwaves in Indochina and for the 5 historical time periods considered. Variable t\_200hPa (temperature at 200hPa) is highlighted with red colors across 5 historical time periods, while the other variables are displayed in gray.}
    \label{fig:relevance_pos_region2}
\end{figure}

\begin{figure}
    \centering
    \includegraphics[width=1.0\linewidth]{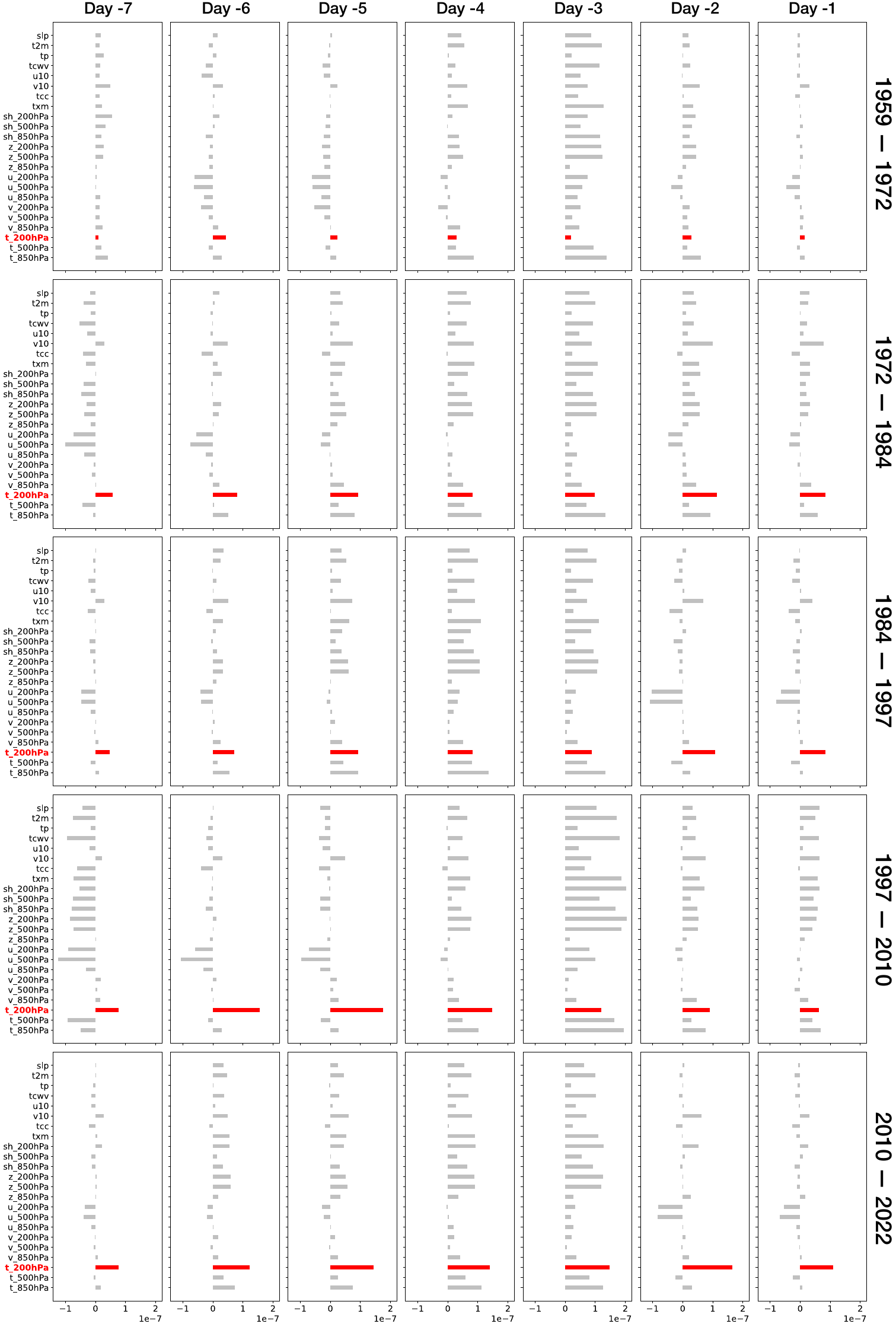}
    \caption{Mean relevance, \textbf{positive and negative}, on \textbf{region 1 + region 2}, for all 23 variables considered, on the 7 days prior to heatwaves in Indochina and for the 5 historical time periods considered.}
    \label{fig:relevance_region12}
\end{figure}

\begin{figure}
    \centering
    \includegraphics[width=1.0\linewidth]{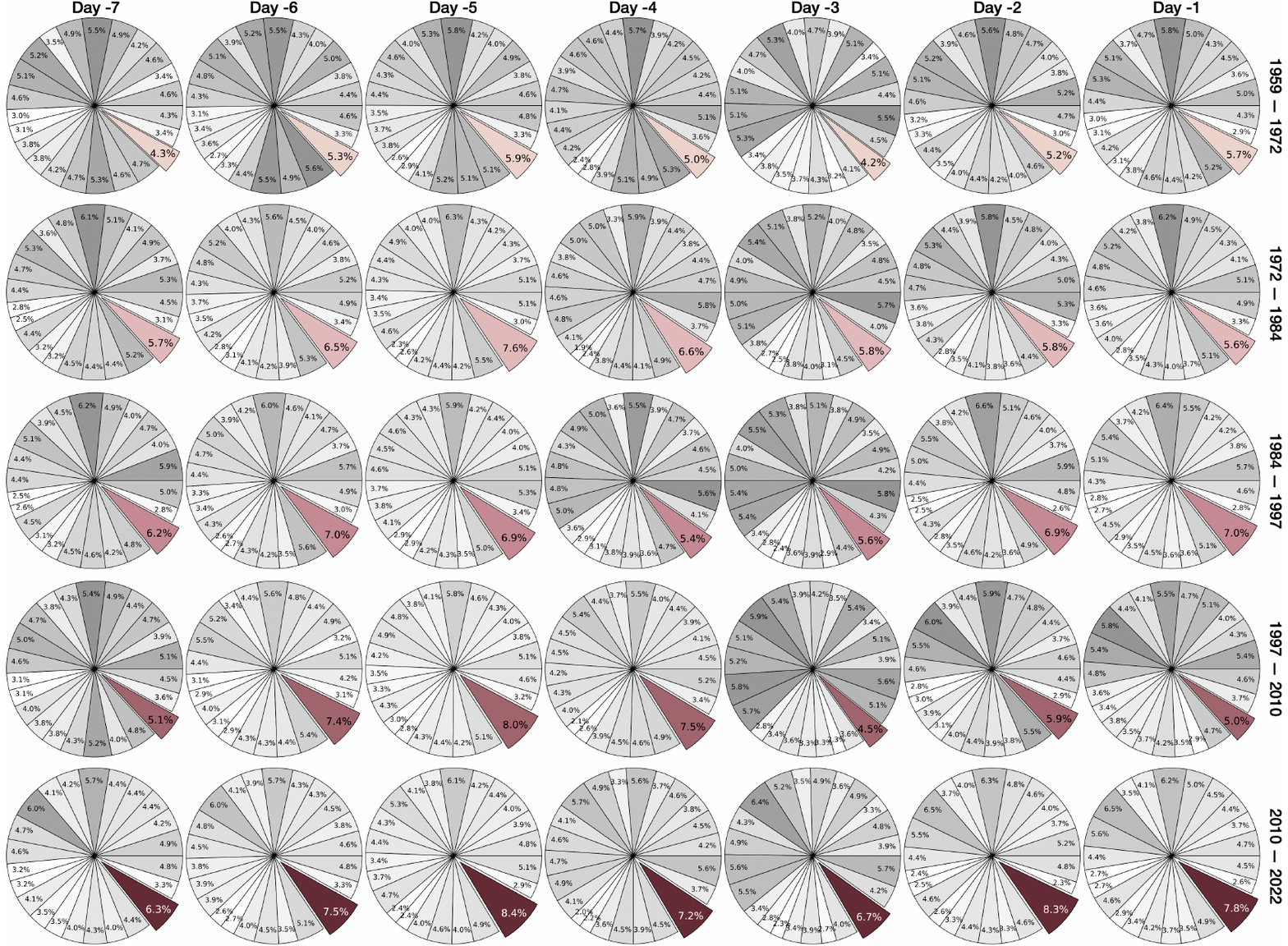}
    \caption{Mean relevance, \textbf{only positive} (negative is set to zero) on \textbf{region 1 + region 2}, for all 23 variables considered, on the 7 days prior to heatwaves in Indochina and for the 5 historical time periods considered. variable t\_200hPa (temperature at 200hPa) is highlighted with red colors across 5 historical time periods, while the other variables are displayed in gray.}
    \label{fig:relevance_pos_region12}
\end{figure}

In this section, we present the mean relevance for all the 23 variables considered, noting how the temperature at 200 hPa is among the variables that is deemed most important by the DL framework for predicting heatwaves in Indochina.
In particular, Figure~\ref{fig:relevance_region1} shows the mean relevance, including positive and negative, for the 23 variables considered in this work, on the 7 days prior to heatwaves, and across the 5 time periods taken into account, for region 1. 
Figure~\ref{fig:relevance_pos_region1} is equal to Figure~\ref{fig:relevance_region1}, except that it only considers positive relevance.
Figures~\ref{fig:relevance_region2}, \ref{fig:relevance_pos_region2}, and Figures~\ref{fig:relevance_region12}, \ref{fig:relevance_pos_region12} show the same quantities as Figures~\ref{fig:relevance_region1}, \ref{fig:relevance_pos_region1} for region 2, and for region 1 and region 2 together, respectively. 


\newpage
\subsection{Relevance vs composite anomalies}
\label{app:comparison}

\begin{figure}[H]
    \centering
    \includegraphics[width=0.8\linewidth]{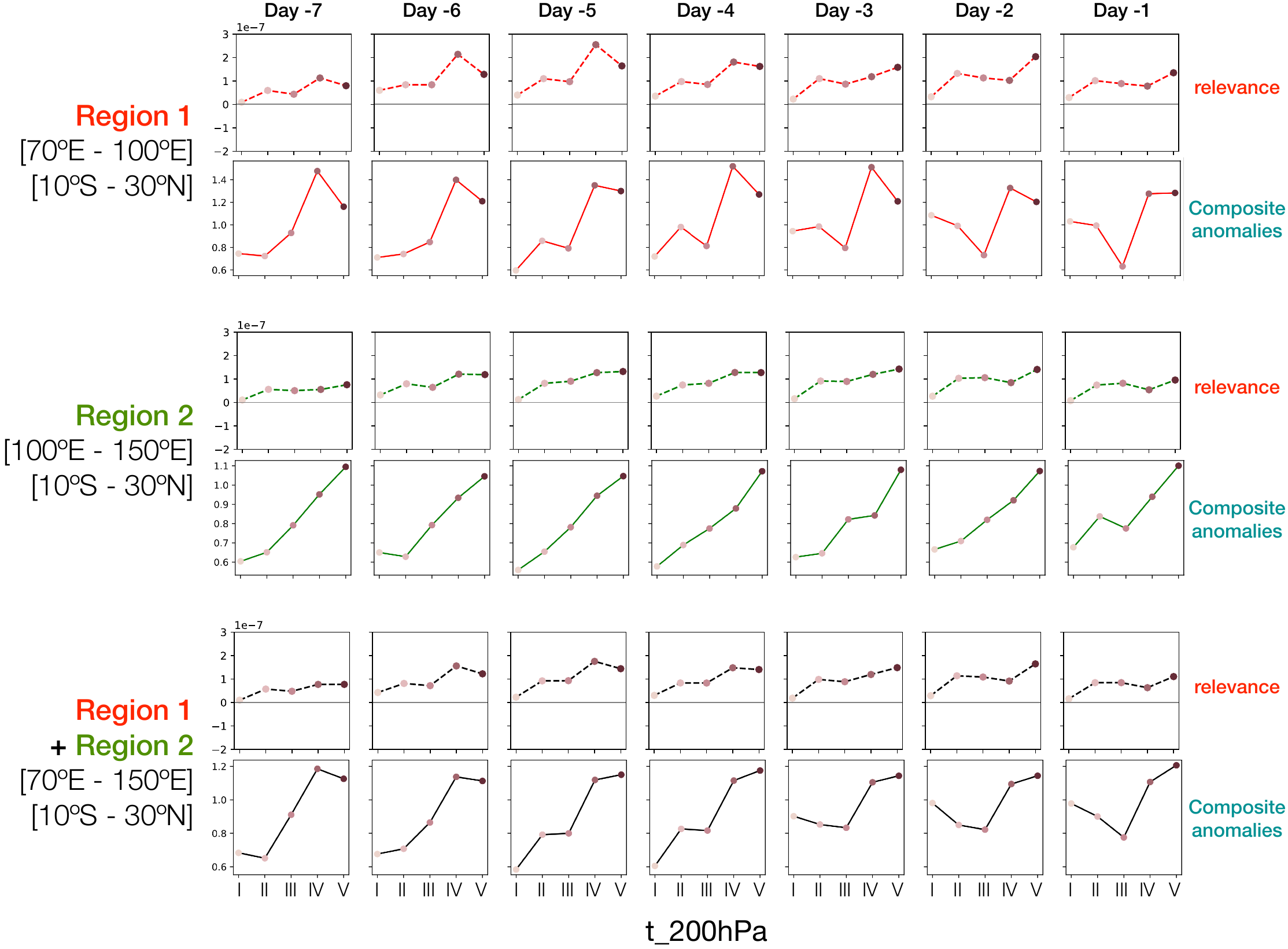}
    \caption{Mean trend of relevance maps vs composite anomalies for variable t\_200hPa (temperature at 200hPa) on region 1 (top row), region 2 (middle row), and region 1 and 2 combined (bottom row).}
    \label{fig:rel_ano_t200hPa}
\end{figure}

\begin{figure}[H]
    \centering
    \includegraphics[width=0.8\linewidth]{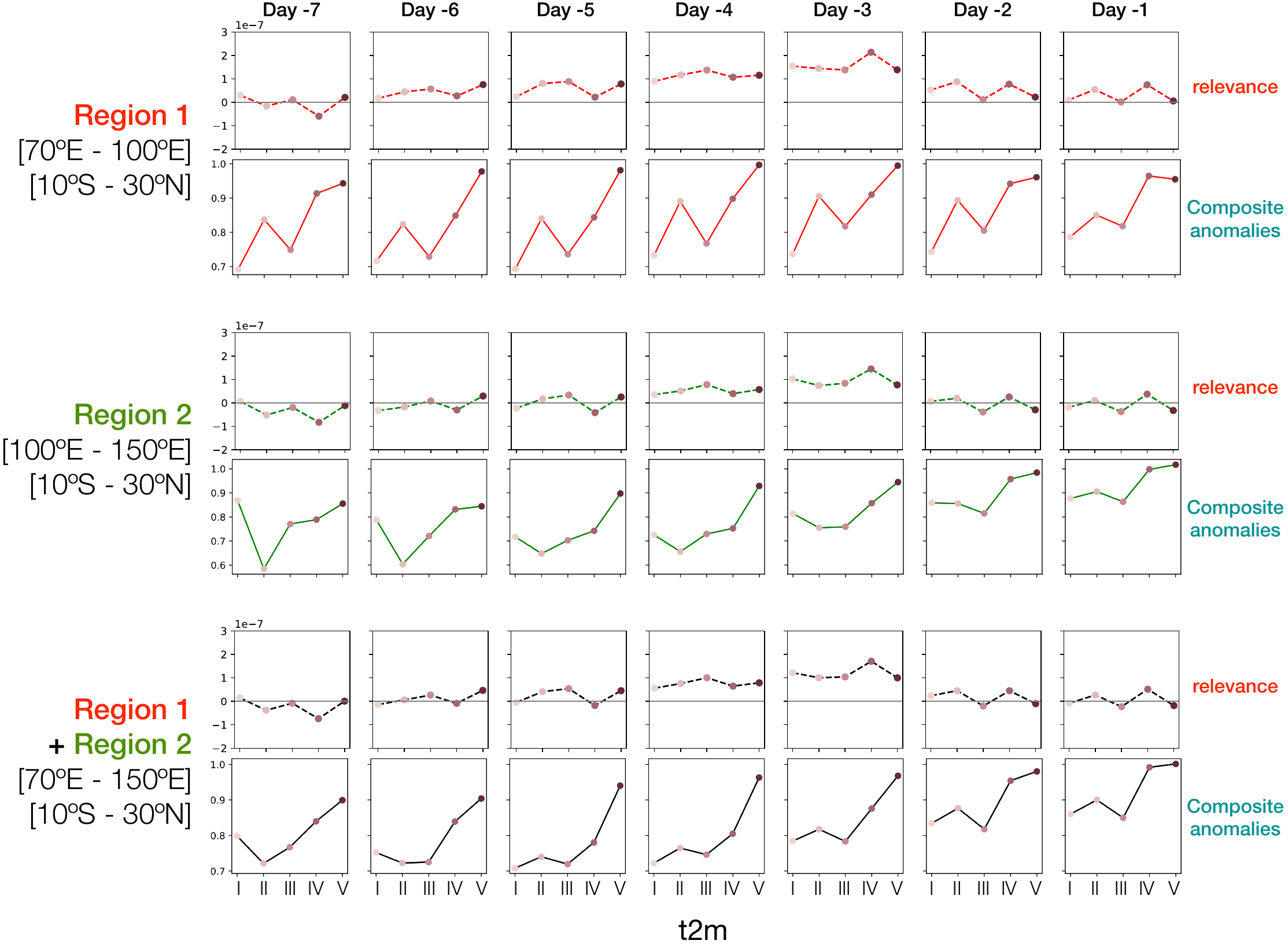}
    \caption{Mean trend of relevance maps vs composite anomalies for variable t2m (2-meter temperature) on region 1 (top row), region 2 (middle row), and region 1 and 2 combined (bottom row).}
    \label{fig:rel_ano_t2m}
\end{figure}

\begin{figure}[H]
    \centering
    \includegraphics[width=0.8\linewidth]{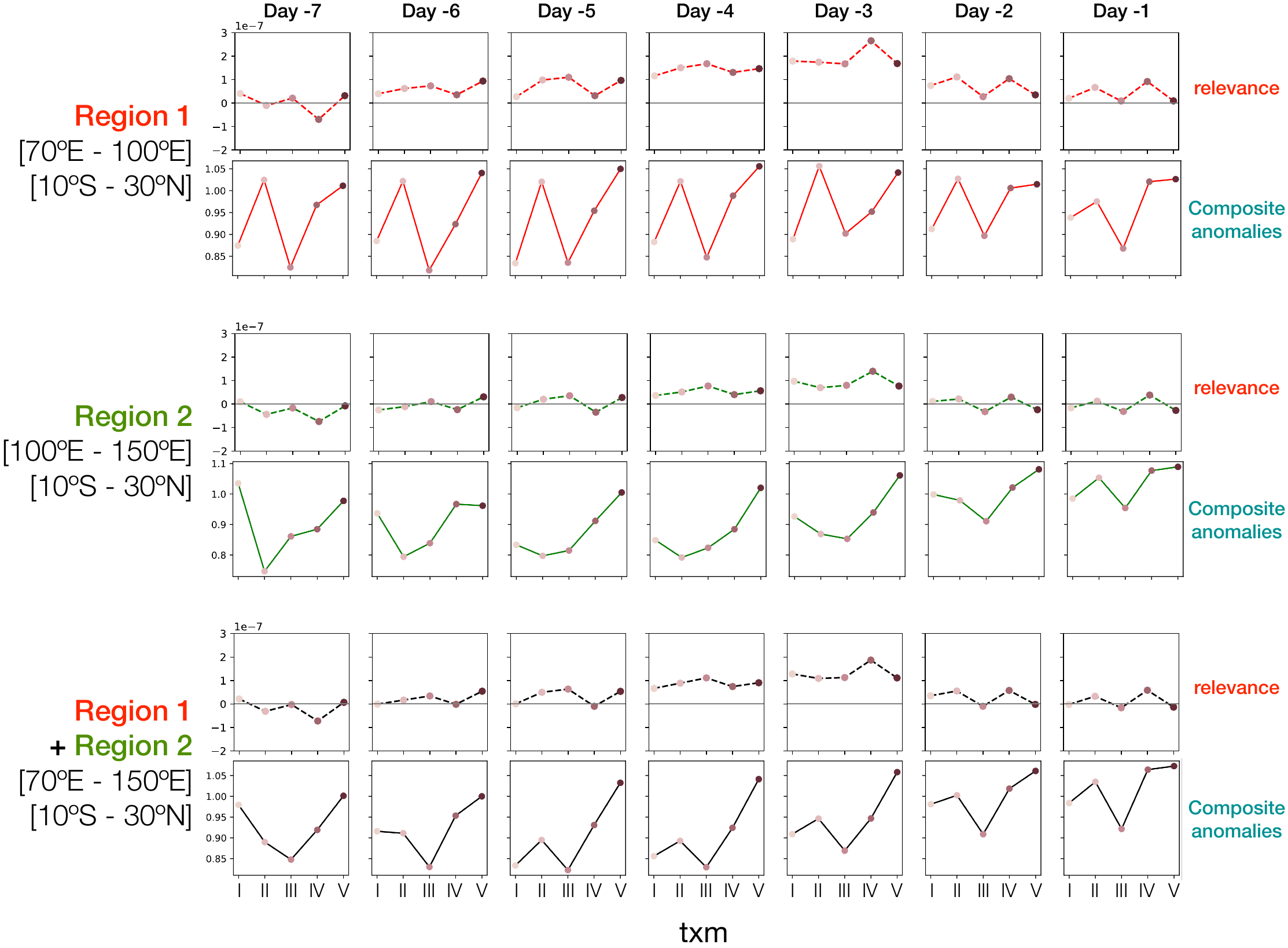}
    \caption{Mean trend of relevance maps vs composite anomalies for variable txm (maximum temperature) on region 1 (top row), region 2 (middle row), and region 1 and 2 combined (bottom row).}
    \label{fig:rel_ano_txm}
\end{figure}

In this section, we present the comparison between the relevance maps and the composite anomalies for thre three regions considered, and for three key variables related to climate change, namely: 200 hPa temperature (Figure~\ref{fig:rel_ano_t200hPa}), 2-meter temperature (Figure~\ref{fig:rel_ano_t2m}), and for maximum temperature (Figure~\ref{fig:rel_ano_txm}).
The results show an upward trend in terms of composite anomalies for all variables and nearly all regions. 
However, relevance exhibits an uptrend only for 200 hPa temperature.

\end{document}

%% file: math_commands.tex
\usepackage{amsmath,amsfonts,bm}

\def\eqref#1{equation~\ref{#1}}
\def\1{\bm{1}}

\DeclareMathAlphabet{\mathsfit}{\encodingdefault}{\sfdefault}{m}{sl}
\SetMathAlphabet{\mathsfit}{bold}{\encodingdefault}{\sfdefault}{bx}{n}